\documentclass[letterpaper, 10 pt, conference]{ieeeconf}  

\IEEEoverridecommandlockouts                              

\usepackage{subcaption}
\usepackage{chemfig}
\setchemfig{atom sep=1.6em}
\usepackage{array}
 
\usepackage[version=4]{mhchem}
\usepackage{listings}
\usepackage{multirow}
\usepackage[table,dvipsnames]{xcolor}
\usepackage[normalem]{ulem} 
\usepackage{pifont}
\usepackage{makecell}
\usepackage{hyperref}

\newcommand{\kefi}[1]{{ \color{black}#1}}
\newcommand{\todo}[1]{{}}
\newcommand{\deleted}[1]{{}}
\newcommand{\tick}{\color{ForestGreen}\pmb{\ding{51}}}
\newcommand{\cross}{\color{red}\pmb{\ding{55}}}

\title{
TARMAC: A Taxonomy for Robot Manipulation in Chemistry
}

\author{Kefeng Huang$^{1}$, Jonathon Pipe$^{1}$, Alice E. Martin$^{2}$, Tianyuan Wang$^{1}$, Barnabas A. Franklin$^{2}$, \\ Andy M. Tyrrell$^{1}$, Ian J. S. Fairlamb$^{2}$, Jihong Zhu$^{1}$ 
\thanks{*This work was supported by the Centre for Doctoral Training in Autonomous Robotic Systems for Laboratory Experiments (ALBERT CDT) at University of York.
}
\thanks{$^{1}$These authors are with School of Physics, Engineering and Technology, University of York
        {\tt\small \{smj553, tnh516, tianyuan.wang, andy.tyrrell, jihong.zhu\}@york.ac.uk}}%
\thanks{$^{2}$ These authors are with Department of Chemistry, University of York
        {\tt\small \{alice.martin, barney.franklin, ian.fairlamb\}@york.ac.uk}}%
}
\begin{document}

\maketitle
\thispagestyle{empty}
\pagestyle{empty}

\begin{figure*}
    \centering
    \includegraphics[width=\textwidth]{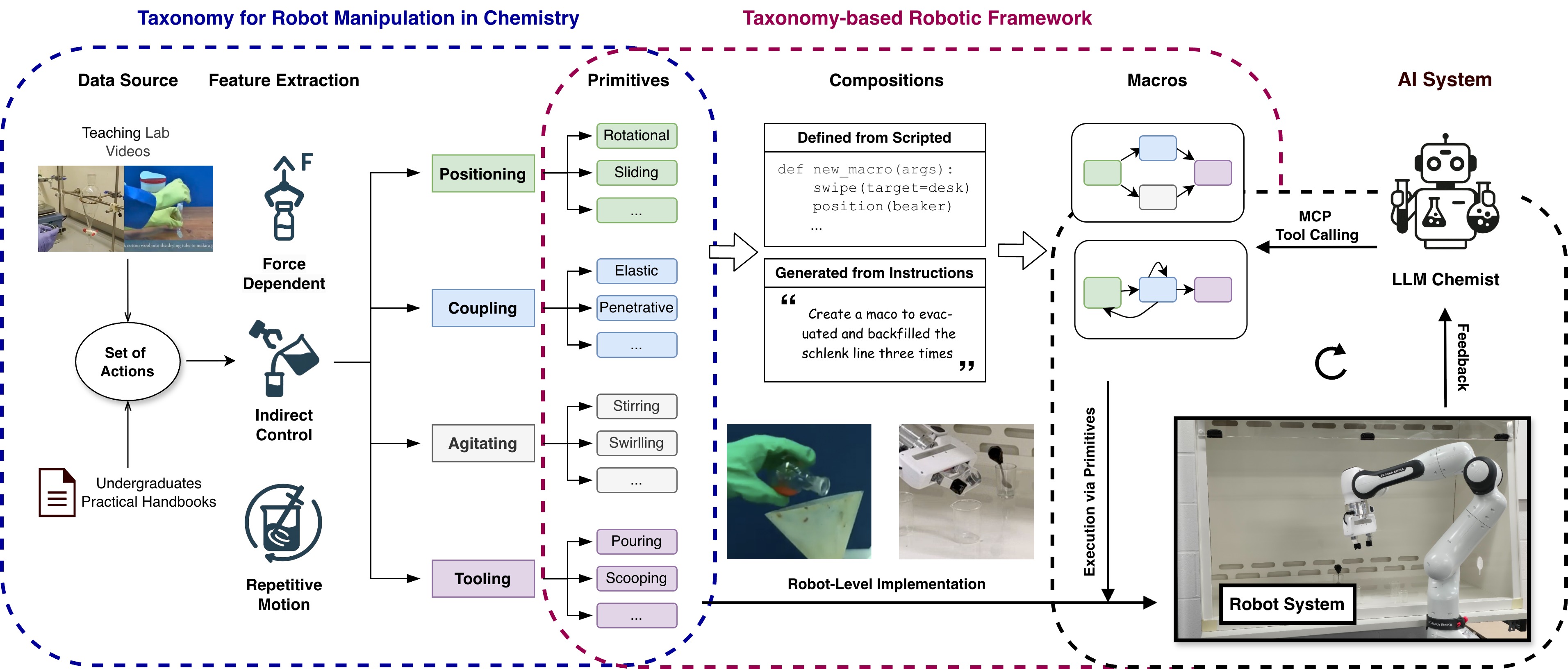}
    \caption{Overview of the TARMAC framework. Laboratory manipulations are first derived from annotated teaching-lab demonstration vidoes and analyzed through feature extraction and experimental validation. These actions are then organized into the TARMAC taxonomy, which provides a structured vocabulary of primitives. The primitives can be directly executed by a robotic platform or composed into higher-level macros, which are exposed through a standardized interface (e.g., \kefi{Model Context Protocol}). This pipeline enables natural-language instructions descriptions to be translated into robot-executable workflows, thereby bridging human intent and robotic capability in the laboratory.}
    \label{fig:tarmac}
\end{figure*}

\begin{abstract}
    Chemistry laboratory automation seeks to increase throughput, reproducibility, and safety, yet many existing systems still rely on frequent human intervention. Recent advances in robotics have \kefi{reduced such dependency}, but without a structured representation of \kefi{the required} skills, their autonomy remains  \kefi{constrained to bespoke, task-specific solutions with limited capacity to transfer beyond their initial design. Existing experiment abstractions primarily encode protocol-level steps, without specifying the robotic actions needed to carry them out. This gap reflects the absence of a systematic account of the manipulation skills required for robots in chemistry laboratories. To fill the critical gap, this work} introduces \textbf{TARMAC} -- a \underline{TA}xonomy for \underline{R}obot \underline{MA}nipulation in \underline{C}hemistry -- a domain-specific taxonomy that defines and organizes the core manipulations required in chemistry laboratories. Using annotated teaching-lab demonstration and supported by experimental validation, TARMAC categorizes actions according to their functional role and physical execution requirements. In addition to serving as a descriptive vocabulary, TARMAC can be instantiated as a robot-executable \textit{primitives} and composed into higher-level \textit{macros}, enabling the reuse of skills and supporting scalable integration into long-horizon workflows. \kefi{These contributions establish a structured foundation for more flexible and autonomous laboratory automation. More information can be found at \url{https://tarmac-paper.github.io/}}
\end{abstract}

\section{Introduction}
Chemistry laboratory automation has been a central pursuit in science and engineering for decades, motivated by the need to accelerate discovery, reduce human workload, improve reproducibility, and enhance operator safety. During this period, a wide range of customized systems and commercial platforms have been developed, each addressing specific experimental workflows or domains. However, despite their growing sophistication, many of these platforms still rely on frequent human intervention -- particularly for tasks that demand dexterity, contextual awareness, or fine manipulation -- limiting the extent to which experiments can be performed fully autonomously.

Recent advances in robotics offer new opportunities to overcome the aforementioned limitations \cite{zhu2025riserobochemist}. General-purpose robotic platforms, including mobile manipulators, have been demonstrated to execute laboratory workflows with minimal supervision \cite{burger2020mobile}. Beyond mobility, increasing attention has turned to the manipulation capabilities of these systems, as the essence of experimental practice lies in the ability to handle objects, tools, and materials with a high degree of precision. Studies utilizing robotic arms to operate vials and prepare samples \cite{lunt2024modular} demonstrate how manipulators can take over tasks that would otherwise require human intervention in existing automation platforms. Yet achieving true autonomy requires expanding beyond the execution of pre-arranged workflows to address a wide range of dexterous operations that current systems cannot perform reliably. These include not only the initial setup of experiments -- such as preparing reagents, calibrating instruments, or connecting fluidic pathways -- but also other skill-intensive tasks throughout experimentation \cite{angelopoulos2024transforming}.  This reveals a critical gap in establishing a well-defined set of robotic skills that can comprehensively cover the requirements of laboratory practice.

A key challenge lies in how such skills are defined, organized, and made accessible to both robotic systems and human researchers, identifying current capability and outline challenges and outlooks. Recent perspectives on laboratory automation emphasize that modular architectures are essential to support interoperability and reusability of experimental procedures \cite{cooper2025accelerating}. Existing digital frameworks, such as chemistry descriptive language ($\chi$DL) \cite{mehr2020universal}, capture high-level intent of how experiments are performed but remain agnostic to the physical execution of actions. They describe what to do, but not how the underlying manipulations should be performed -- an abstraction that facilitates hardware-agnostic protocol design but leaves the concrete requirements of robotic execution unspecified. Bridging this gap demands a structured vocabulary of laboratory actions —- one that reduces ambiguity, supports systematic reuse of skills, and enables robots to interpret, plan, and execute experiments with fidelity.

This work introduces TARMAC -- a \underline{TA}xonomy for \underline{R}obot \underline{MA}nipulation in \underline{C}hemistry -- designed to categorize and describe the core manipulations required in laboratory practice. \kefi{As illustrated in Figure~\ref{fig:tarmac}, TARMAC is }derived from annotated demonstrations from teaching-lab videos and experiment handbooks with experimental validation \kefi{and developed through a joint effort between chemists and roboticists, combining knowledge from both domains to} organize laboratory actions into categories that reflect both their functional role in experiments and their physical execution requirements. Beyond serving as a structured vocabulary, TARMAC can be instantiated as atomic action primitives, which can then be composed into higher-level workflows or integrated with large language models (LLMs) for planning and execution. In this way, it bridges the gap between protocol-level descriptions familiar to chemists and the robot-executable skills required for automation.

The contributions of this work can be summarized as follows:
\begin{itemize}
\item A domain-specific taxonomy (TARMAC) that systematizes manipulation skills in chemistry.
\item An annotated dataset of laboratory manipulations supporting taxonomy development and evaluation.
\item A proof-of-concept framework showing how TARMAC can ground action primitives and enable LLM-driven translation of instructions into robotic execution.
\end{itemize}

The remainder of this paper is structured as follows. Section~\ref{sec:related_works} reviews previous work in laboratory automation and robotic manipulation taxonomies. Section~\ref{sec:methodology} outlines the methodology used to derive the proposed taxonomy, which is then presented in detail in Section~\ref{sec:taxonomy}. Section~\ref{sec:analysis} provides an analysis of the taxonomy, while Section~\ref{sec:validation} demonstrates its practical validation through a framework applied to representative laboratory tasks. Finally, Section~\ref{sec:challenges} discusses the key challenges and future outlook, and Section~\ref{sec:conclusion} concludes the paper.

\section{Related Works} \label{sec:related_works}
This section reviews three strands of prior work that inform the development of the taxonomy. The first concerns laboratory automation systems, particularly those incorporating robotic manipulators, which illustrate the state of the art in executing chemistry experiments and demonstrate how procedures can be abstracted into executable steps. The second strand examines existing manipulation taxonomies, which formalize action spaces through structured vocabularies in other domains and thereby highlight the need for a chemistry-specific taxonomy. The third strand considers research on large language models (LLMs) in robotics, where advances in long-horizon planning and reasoning motivate the instantiation of the taxonomy as a bridge between abstract instructions and executable skills, reinforcing the need for a well-defined action space.

\subsection{Lab Automation} 
Automation in chemistry has been a longstanding pursuit, with origins tracing back to early laboratory devices and later formalized through structured methodologies such as Design of Experiments (DoE) and High-Throughput Experimentation (HTE) \cite{dar2004high}. These approaches established that the value of automation lies not only in mechanical execution, but also in systematic abstraction, allowing experiments to be scaled, replicated, and optimized efficiently. This foundation set the stage for increasingly sophisticated forms of laboratory autonomy.

Recent advances in robotics have emerged as a promising direction for pushing automation further \cite{zhu2025riserobochemist}. Robotic systems bring the ability to manipulate diverse laboratory tools and materials directly, thereby reducing human intervention and enabling higher levels of autonomy in experimental workflows \cite{angelopoulos2024transforming}. Landmark studies employing mobile manipulators have shown that autonomous platforms can coordinate distributed laboratory stations to execute complete workflows with minimal human intervention \cite{burger2020mobile}, \kefi{and more recently, have been extended to exploratory synthetic chemistry through the integration of diverse analytical modalities and heuristic decision-making \cite{dai2024autonomous}.} Beyond this opening, increasing attention has turned to the manipulation capabilities of robots themselves. \kefi{Subsequent efforts have extended these workflows by incorporating robotic manipulators to replace specific stations, enabling tasks such as vial opening and sample preparation \cite{lunt2024modular}. Beyond such integrated systems, robotic arms have also been demonstrated to perform more dexterous laboratory skills in isolation,} such as pouring \cite{yoshikawa2022chemistry} and scraping\cite{pizzuto2024accelerating}, indicating that general-purpose platforms can increasingly take on operations once restricted to human chemists. These works underscore that dexterous manipulation, rather than mobility alone, is another key to advancing laboratory autonomy.

Recent discussions in laboratory automation have converged on the view that modularity is a key requirement for achieving robustness, interoperability, and reusability across systems \cite{cooper2025accelerating}. Much of this modularity has so far been realized at the level of station-based workflows, with each step executed by a dedicated platform. However, such a station-based perspective often neglects the general-purpose potential of robotic manipulators, treating them as substitutes for predefined modules rather than as versatile agents capable of performing diverse actions. In parallel, digital frameworks such as $\chi$DL \cite{mehr2020universal} have sought to abstract experimental procedures into machine-readable formats, enabling portability and reproducibility across platforms. These abstractions are well suited for describing chemical workflows, but they primarily capture the sequence of procedures without detailing how the underlying manipulations are performed. \kefi{For example, the same $\chi$DL “addition” step can be carried out by a syringe pump on a Chemputer, by a pipetting head on an Opentrons robot, or by a Kinova robotic arm \cite{rauschen2024universal}, but the abstraction does not specify which modality or scale is intended.} The lack of such specification results in general-purpose robots being constrained to the programming of isolated procedures, necessitating repetitive and task-specific development for each new experimental protocol. Addressing this gap requires a structured taxonomy of laboratory manipulations —- one that explicitly defines the repertoire of skills underpinning experimental practice and enables their reuse as abstractions across diverse workflows.

\subsection{Manipulation Taxonomies}

There has been extensive work on taxonomies of robotic manipulation, which generally serve two main purposes. First, they act as a descriptive language that organizes complex actions into structured categories, thereby reducing ambiguity and deepening understanding of manipulation problems. For instance, taxonomies for deformable object handling \cite{blanco2024t} classify actions by deformation modes, energy regimes, and interaction patterns, capturing the distinctive challenges of non-rigid materials. Second, taxonomies can inform the design of future systems, shaping both system development and hardware design. An example is the abstraction of human grasping into a compact set of primitives with modifiers \cite{heinemann2015taxonomy}, which has guided the design of grasp controllers and robotic hands. At a broader scope, field-wide schemes \cite{paulius2019manipulation} highlight how taxonomies can encode skills in machine-readable form, unifying heterogeneous task descriptions and facilitating transfer across applications.

\kefi{Building on these perspectives, a domain-specific taxonomy tailored to chemistry laboratories is developed. The aim is not only to catalogue the diverse manipulations required in experimental practice, but also to identify the fundamental challenges they present and distill them into a structured abstraction. Such an abstraction provides a reusable foundation that can inform the design of laboratory automation systems, guide the specification of robotic capabilities, and promote consistency across heterogeneous workflows.}

\subsection{LLMs in Robotics and Chemistry}
Large language models have recently attracted growing interest in robotics due to their ability to map natural language into structured instructions and to support general reasoning. Early work demonstrated that LLMs can be prompted to generate executable code, enabling robots to perform complex behaviors without task-specific training \cite{liang2022code}. This “code-as-policy” approach illustrates how foundation models can act as program synthesizers, translating human intent into robot-executable routines \kefi{, but it operates in an open-loop manner, with no mechanism to regenerate or adapt code in response to runtime feedback. Subsequent work advanced this paradigm by introducing a closed-loop structure, in which LLM-generated action proposals are continually evaluated against an affordance-based value function \cite{huang2022inner}, ensuring that executed behaviors remain consistent with the robot’s capabilities and environmental constraints. Nevertheless, across both approaches, the range of achievable behaviors remains ultimately bounded by the scope of the underlying perception and control APIs.}

\kefi{
In chemistry automation, previous works have shown the potential of large language models to translate natural-language protocols into machine-usable forms \cite{ruan2024automatic}. For example, natural-language procedures can be expressed in the $\chi$DL domain-specific language (DSL) and iteratively refined through verifier-assisted prompting to ensure syntactic and constraint validity \cite{yoshikawa2023large}, while more recent systems such as ORGANA demonstrate that LLM-guided assistants can move beyond protocol translation to plan, schedule, and execute multi-step laboratory experiments \cite{darvish2025organa}. Despite this promise, such approaches highlight the continuing need for verification and structural alignment: LLMs are not inherently constrained by the grammar or semantics of DSLs like $\chi$DL, and $\chi$DL itself was designed as a workflow abstraction for automation pipelines rather than as a direct specification of robot-level actions. A similar issue arises more broadly in robotics, where LLM-generated plans must still be grounded in diverse perception and control APIs that are typically hand-integrated. To reduce this reliance on ad-hoc interfaces, the Model Context Protocol (MCP) \cite{anthropicmcp} offers a standardized interface for exposing robot skills as callable functions, thereby providing a more systematic path to ground high-level instructions in executable actions and unifying robot APIs within a common framework, for instance by connecting chemistry-centric descriptions like $\chi$DL with robot-executable primitives. 

Together, these efforts highlight the importance of both structured action representations and standardized interfaces. This taxonomy extends this direction by providing a comprehensive repertoire of chemistry-specific skills that can be instantiated as callable functions within an MCP framework, thereby ensuring both broad coverage of laboratory actions required and ease of integration with LLM-based planning systems.
}

\section{Methodology} \label{sec:methodology}
To develop a meaningful and generalizable taxonomy of robotic manipulations, a systematic process was established to identify and justify the core features underpinning such tasks. The process began with the collection of representative video data capturing a diverse range of laboratory manipulations. In collaboration with chemists, the raw demonstrations were segmented and annotated to isolate individual actions, ensuring that the decomposition reflected authentic laboratory practice. These segmented actions enabled the analysis of recurring patterns of interaction, from which a set of candidate features fundamental to describing manipulation behaviors was extracted. The importance of these features was then validated through experimental measurements, ensuring that they were not arbitrarily selected but instead reflected essential aspects of the tasks. The following subsections detail this pipeline, from data collection to feature validation.

\subsection{Data Sources and Collections}

The dataset was compiled from a series of video demonstrations produced by the Chemistry Teaching Laboratories. These videos are used to provide practical training for undergraduate students in Chemistry, Biochemistry, and Natural Sciences, and constitute a systematic instruction in laboratory manipulations offered during their degree programmes. This material is representative for the present study, as the skills taught to students capture fundamental and widely applicable techniques of laboratory practice. In total, 91 videos were collected, amounting to 273 minutes of footage. This dataset provides a foundation for systematically reviewing the core features of chemical manipulation tasks.

\subsection{Task Segmentation and Annotation}

\begin{figure}[t]
    \begin{subfigure}{0.48\linewidth}
        \centering
        \includegraphics[width=\linewidth]{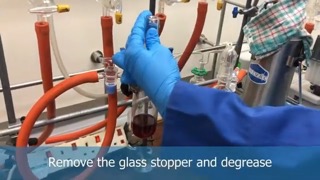}
        Pull glass stopper\\
        \texttt{linear coupling}
    \end{subfigure}
    \begin{subfigure}{0.48\linewidth}
        \centering
        \includegraphics[width=\linewidth]{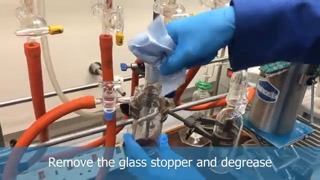}
        Wipe grease\\
        \texttt{wiping}
    \end{subfigure}
    \vspace{0.5em}
    \newline
    \begin{subfigure}{0.48\linewidth}
        \centering
        \includegraphics[width=\linewidth]{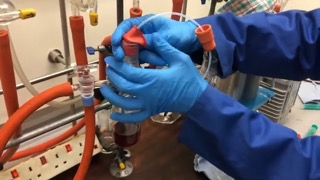}
        Insert cannula stopper \\
        \texttt{elastic\\coupling}
    \end{subfigure}
    \begin{subfigure}{0.48\linewidth}
        \centering
        \includegraphics[width=\linewidth]{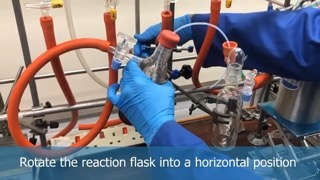}
        Rotate flask \\
        \texttt{rotational\\positioning}
    \end{subfigure}
    \caption{Examples of annotated frames from the chemistry teaching videos. Each frame shows a representative laboratory manipulation alongside its natural-language description (top) and \kefi{later} corresponding taxonomic actions(bottom).}
    \label{fig:chemistry_manipulation_dataset}
\end{figure}

Each teaching video was manually reviewed and segmented into discrete manipulation steps through joint effort between roboticist and chemists. Segmentation was carried out at the level of atomic actions (e.g., pouring a liquid, adjusting a pipette), defined as the smallest meaningful unit of laboratory interaction. Annotators then labelled each segment using intuitive descriptions of the observed action. Importantly, the labellers were not restricted to a predefined action set; instead, labels were allowed to emerge directly from the data, capturing the full diversity of manipulations present in the demonstrations. A few examples of the annotated frames and labels are shown in Figure~\ref{fig:chemistry_manipulation_dataset}. In total, 563 instances of individual action segments were identified and annotated across the dataset. The resulting set of raw labels was subsequently refined through an iterative consolidation process. Redundant or synonymous labels were merged, and closely related variants were grouped into broader categories. This process yielded a reduced vocabulary of  distinct manipulation types, which served as the basis \kefi{to find some common characteristics}. 

\subsection{Feature Extraction} \label{sec:features}

To progress from descriptive action labels toward a principled taxonomy, three core features were identified as consistently underlying laboratory manipulations. These features, derived from recurring patterns observed across the annotated demonstrations, capture aspects of control that are critical to task execution:

\subsubsection{Force Dependence}
Many manipulations are defined not solely by geometric trajectories but also by the wrench signals (forces and torques) that arise during execution. Importantly, such forces may need to be applied even in the absence of motion. For example, inserting a stopper requires maintaining axial force at the final stage to achieve a proper seal, even once the stopper has stopped moving. Similarly, rotating a clamp screw relies on applying torque until sufficient resistance is reached. In both cases, wrench feedback governs the termination condition and ensures successful task completion.

\subsubsection{Motion Pattern}
Most manipulations can be characterized as either one-off motions or inherently repetitive/cyclic motions. A key difference lies in whether the action could in principle be completed in a single execution or whether repetition is unavoidable. For example, pouring liquid until a target volume is reached may be performed as a continuous one-off motion (holding the container steadily) or as a repetitive sequence (tilting in small increments until the desired level is achieved). This makes pouring a one-off action, and each increment is itself a complete execution. By contrast, actions such as shaking or stirring cannot naturally be executed in a single motion; they require repeated cycles until a condition is satisfied (e.g., the solution becomes homogeneous), which is why stirring is classified as cyclic. 

\subsubsection{Control Directness}
Manipulations vary in how directly the operator (or robot) interacts with the target object. In some cases, control is direct -- for example, grasping and lifting a vial acts immediately on the object of interest. In other cases, control is indirect, mediated through another object: tilting a flask to pour liquid affects the liquid only via the container. This distinction reflects the degree of coupling between the \kefi{motion} and the outcome. Indirect actions often require additional care in planning and execution compared to direct ones.

Together, these three features provide a compact yet expressive basis for describing manipulation behaviors, serving as the key to distinguish for the taxonomy presented in the following section.

\subsection{\kefi{Force Dependency} Validation}

While motion pattern and control directness are apparent from visual observation and readily understood in intuitive terms, the role of force dependence required explicit validation. A series of experiments was therefore conducted to measure the forces applied during representative laboratory manipulations. The results confirmed that many actions rely critically on detecting and regulating wrench signals.

\subsubsection{Experiment Setup}

\begin{figure}
    \centering
    \includegraphics[width=\linewidth]{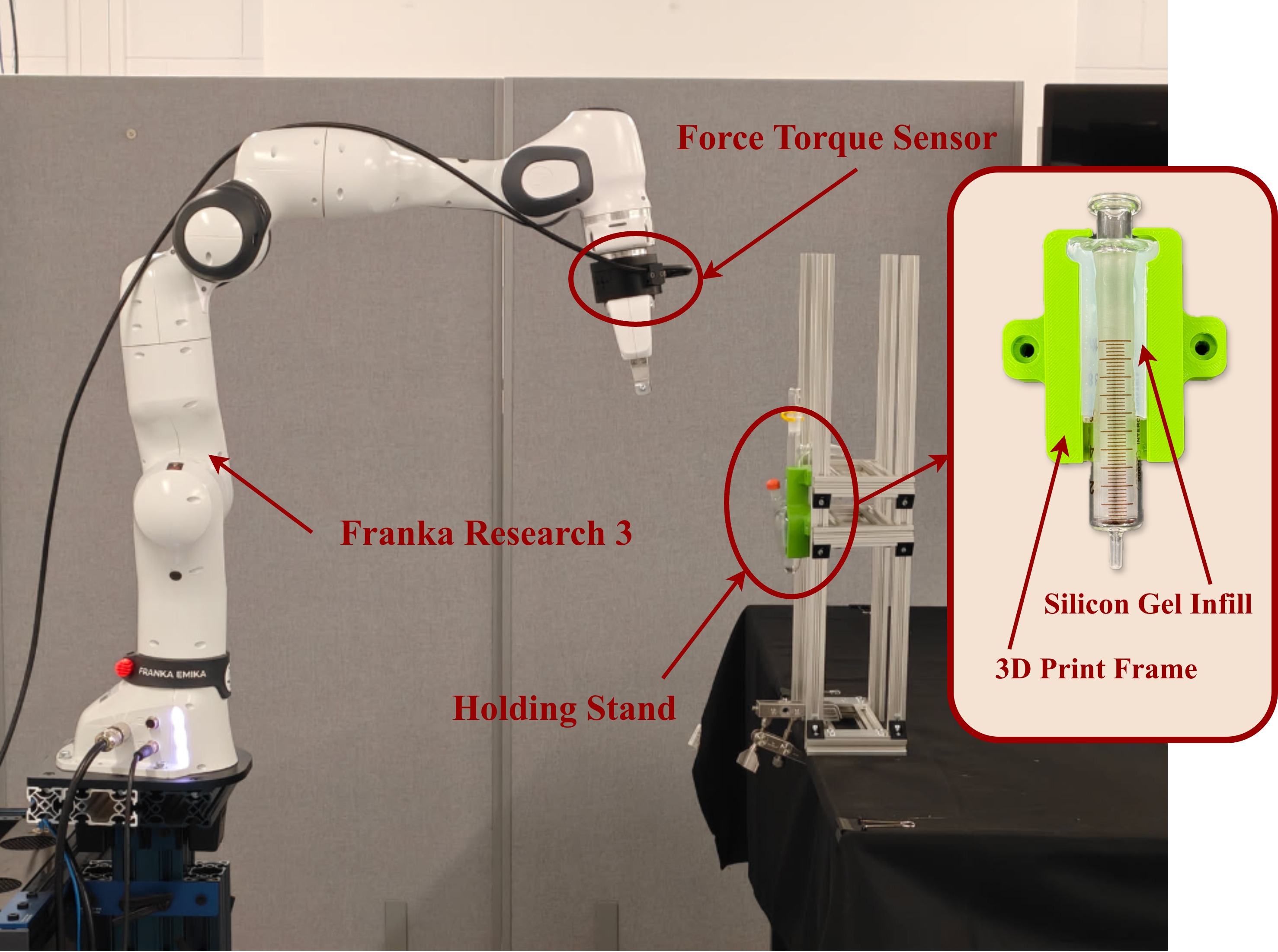}
    \caption{Experimental setup for force–torque measurements. A Franka Research 3 robotic arm with a parallel gripper and a Bota Systems LaxONE sensor was used to manipulate laboratory glassware. The inset highlights the custom 3D-printed fixture with silicone gel padding, mounted on an aluminum frame to securely hold instruments.}
    \label{fig:force_measure_setup}
\end{figure}

The experiments were conducted using a Franka Research 3 robotic arm equipped with a Bota Systems LaxONE six-axis force–torque sensor, mounted between the robot flange and a parallel gripper. To securely hold various laboratory instruments, a custom fixture was designed, consisting of a 3D-printed outer frame with silicone gel padding. The gel not only provided compliance to minimize the risk of breakage but also compensated for small mismatches between the printed geometry and the actual glassware, ensuring a snug fit. The fixture was mounted on an aluminum frame fixed to the desk to provide a stable working environment for accurate force measurements. The overall setup is illustrated in Figure~\ref{fig:force_measure_setup}.

\kefi{To examine the role of force dependence in laboratory procedures, a representative set of manipulation tasks was selected, spanning both force-intensive and visually guided actions. The force-dependent group included operations such as pushing a cap onto a vial, tightening a screw cap, and inserting a needle into a rubber septum, each hypothesized to require force feedback for reliable completion. As a point of contrast, tasks expected to rely primarily on visual cues, such as rotating a tap handle, were also incorporated.} Each task was executed by the robot following pre-programmed trajectories, while recording force–torque wrench data, end-effector poses, and RGB images from a side-mounted Intel RealSense D435 camera that captured the manipulation details. All data streams were synchronized and logged at 10 Hz for subsequent analysis.

\begin{figure*}[t]
    \centering
    \begin{minipage}[t]{0.2\textwidth}
        \begin{subfigure}{\linewidth}
            \centering
            \includegraphics[height=8em]{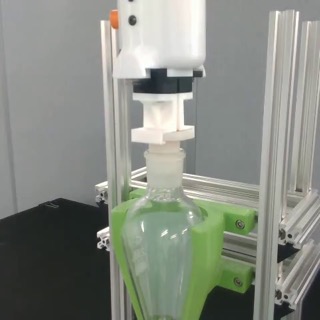}
            \caption*{Pushing Cap}
        \end{subfigure}
        \newline
        \begin{subfigure}{\linewidth}
            \centering
            \includegraphics[height=10em]{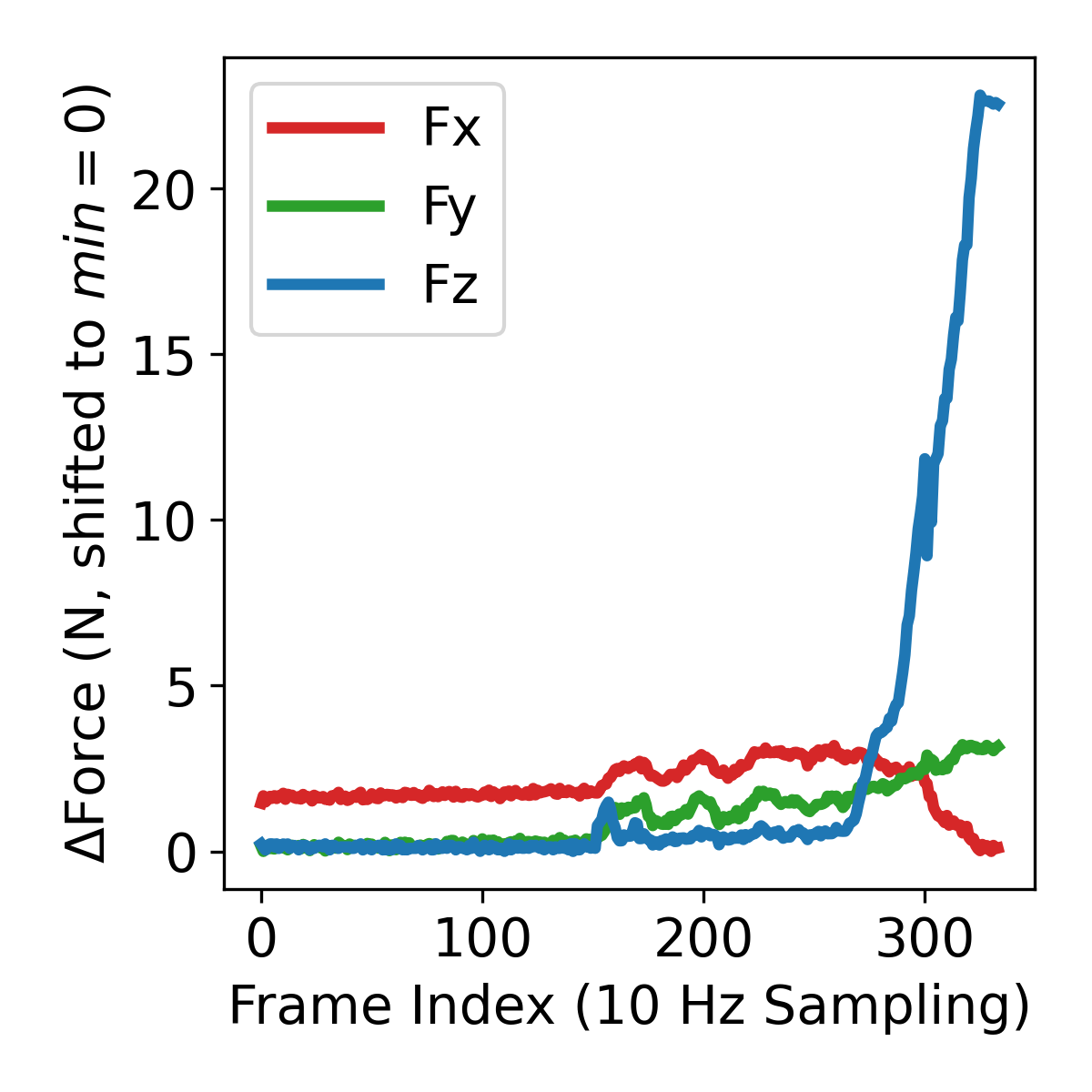}
            \caption{}
            \label{fig:force_pushing}
        \end{subfigure}
    \end{minipage}
    \vline
    \begin{minipage}[t]{0.36\textwidth}
        \hfill
        \begin{subfigure}{0.44\linewidth}
            \centering
            \includegraphics[height=8em]{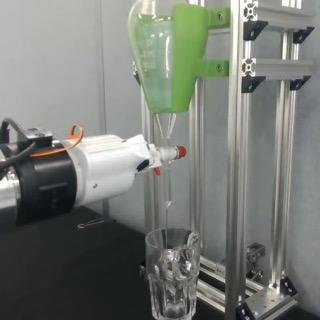}
            \caption*{Rotating Tap}
        \end{subfigure}
        \hfill
        \begin{subfigure}{0.44\linewidth}
            \centering
            \includegraphics[height=8em]{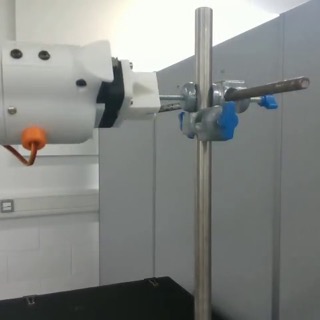}
            \caption*{Tightening Screw}
        \end{subfigure}
        \hfill
        \newline
        \begin{subfigure}{\linewidth}
            \includegraphics[height=10em]{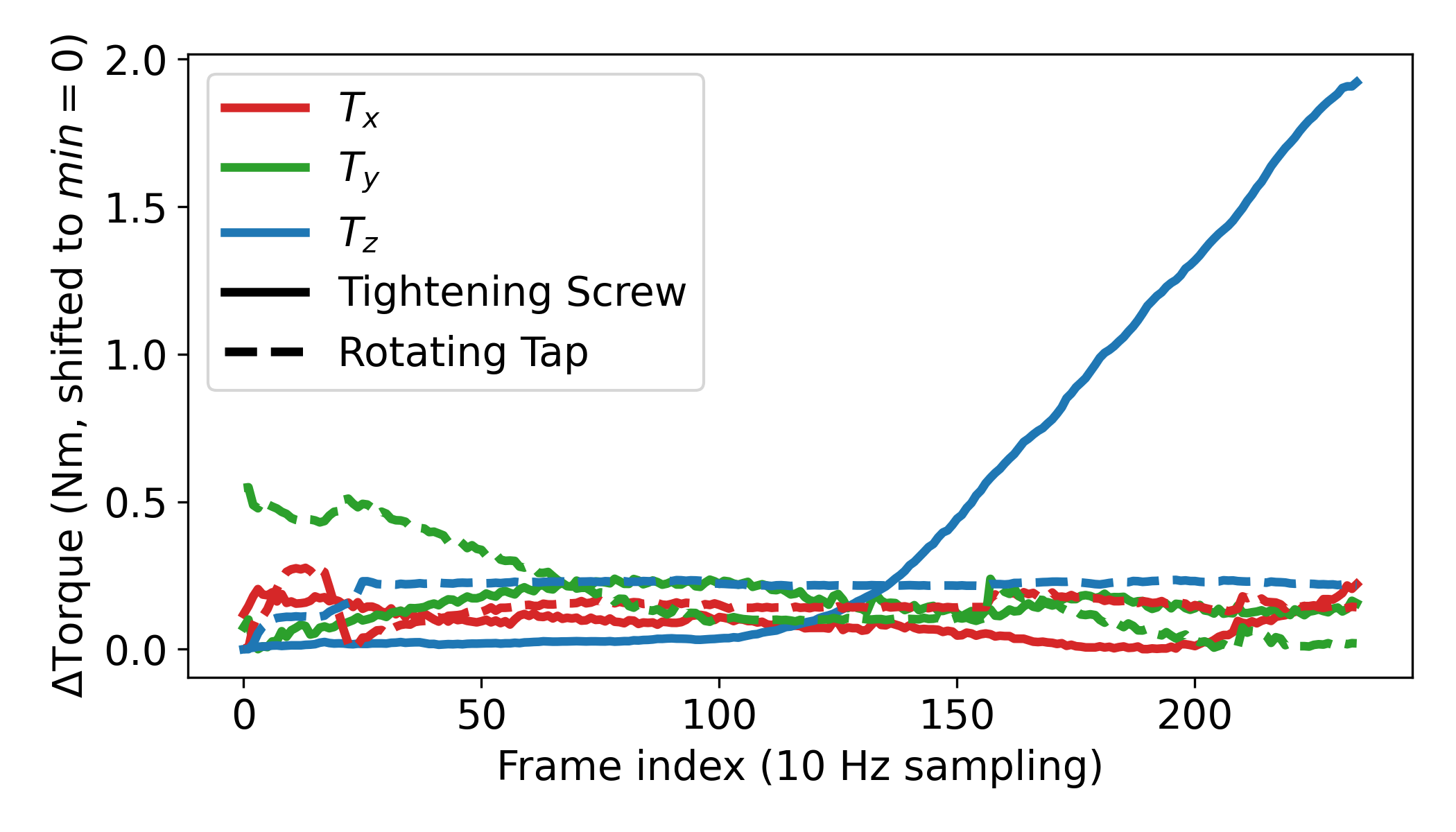}
            \caption{}
            \label{fig:torque_comparison}
        \end{subfigure}
    \end{minipage}
    \vline
    \begin{minipage}[t]{0.36\textwidth}
        \hfill
        \begin{subfigure}{0.44\linewidth}
            \centering
            \includegraphics[height=8em]{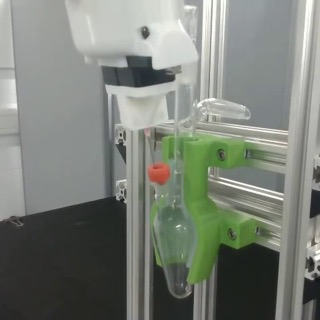}
            \caption*{$t_1=207$}
        \end{subfigure}
        \hfill
        \begin{subfigure}{0.44\linewidth}
            \centering
            \includegraphics[height=8em]{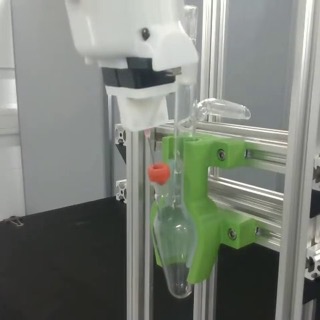}
            \caption*{$t_2=212$}
        \end{subfigure}
        \hfill
        \newline
        \begin{subfigure}{\linewidth}
            \includegraphics[height=10em]{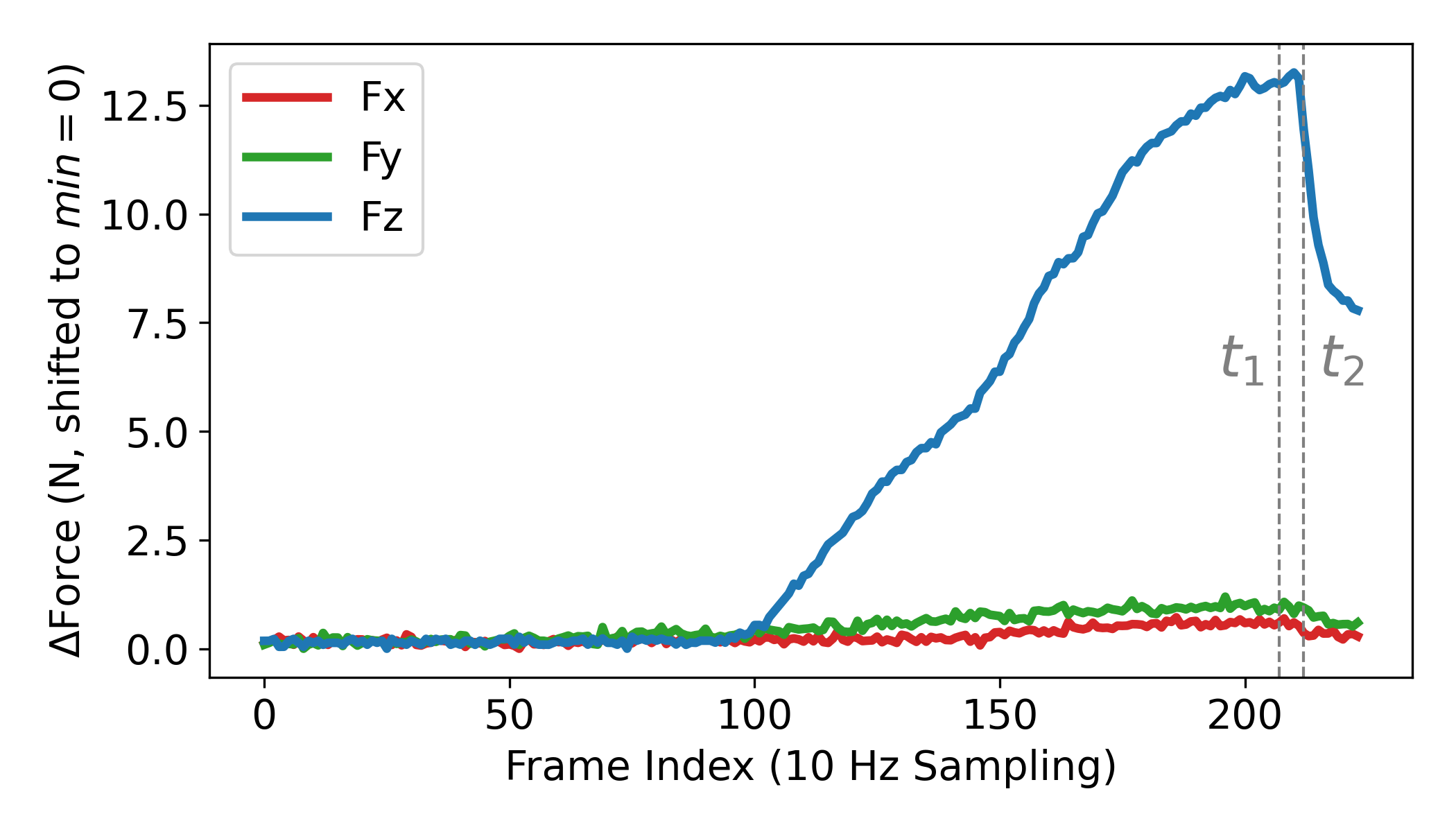}
            \caption{}
            \label{fig:force_insert}
        \end{subfigure}
    \end{minipage}
    \caption{Experimental validation of force dependence in representative laboratory manipulations. 
(a) Cap insertion: \kefi{Red, green, and blue lines denote forces along the $x$ ($F_x$), $y$ ($F_y$), and $z$ ($F_z$) axes.} A sharp peak in $F_z$ provides a termination signal, ensuring a secure seal without glass breakage. 
(b) Rotational comparison: \kefi{Red, green, and blue lines denote torques about the $x$ ($T_x$), $y$ ($T_y$), and $z$ ($T_z$) axes. Solid and dashed curves correspond to clamp screw tightening and tap rotation, respectively.} Tap rotation shows relatively flat torque profiles governed by visual alignment, whereas screw tightening exhibits steadily increasing torque, indicating task completion. 
(c) Needle puncture: \kefi{Red, green, and blue lines denote forces along the $x$ ($F_x$), $y$ ($F_y$), and $z$ ($F_z$) axes. }Insertion into a rubber septum produces a force build-up followed by a sudden drop upon puncture—a transition undetectable by vision alone. 
Together, these results highlight the critical role of force–torque signals in defining reliable stopping conditions for laboratory actions.}
    \label{fig:force_exp_results}
\end{figure*}

\subsubsection{Analysis of Results}

A representative selection of recorded force profiles is shown in Figure \ref{fig:force_exp_results}, with the full dataset provided in the \href{https://tarmac-paper.github.io/}{website}. These results illustrate how force dependence plays a central role in determining termination conditions for laboratory manipulations.

The comparison between tap rotation and clamp-screw tightening (Figure \ref{fig:torque_comparison}) demonstrates that, although the two actions share similar rotational motions, they rely on fundamentally different stopping signals. In tap rotation, the task is governed by visual control: the operator adjusts the angular position to achieve the desired flow rate, and the torque profile remains relatively flat. By contrast, screw tightening is not defined by a visual target but by achieving sufficient resistance to secure the connection. Here, the steadily rising torque around the z-axis provides the termination criterion, ensuring the cap is stable without over-tightening.

Cap insertion (Figure \ref{fig:force_pushing}) demonstrates a similar principle, where termination is defined by reaching a threshold of axial force. In this case, the experimental data show how the force profile rises as the cap is pressed into place, with the threshold marking completion of the seal. While these experiments did not directly measure the consequences of exceeding this threshold, in laboratory practice it also functions as a safety constraint: applying excessive force risks fracturing the glass vial, while insufficient force results in an incomplete seal. Thus, this task exemplifies how a single threshold can simultaneously signal completion and delimit safe operation.

Needle insertion into a rubber septum (Figure \ref{fig:force_insert}) further illustrates the importance of force-based stopping signals. The force increases steadily until a puncture occurs, at which point the profile exhibits a sudden drop (in force). Crucially, this transition cannot be inferred from the visual scene—even at timestamps $t_1$ and $t_2$, immediately before and after puncture, the setup appears almost identical. The force profile therefore provides the only reliable indication of task completion.

Taken together, these results validate that force and torque signals are essential in defining termination conditions for certain laboratory manipulations. While visual feedback may suffice in some cases (e.g., tap rotation), other tasks depend on thresholds or patterns in force or torque to determine completion reliably. 

\section{The Taxonomy} \label{sec:taxonomy}

\begin{table}
    \centering
    \begin{tabular}{|c|c|c|c|c|}
        \hline
        \rowcolor{gray!20} \makecell{Force \\ Dependent} & \makecell{Indirect \\ Control}& \makecell{Repetitive \\ Motion} & Count $\downarrow$ & \makecell{Taxonomic Category}\\
        \hline 
         \cellcolor{yellow!30} \tick & \cross & \cross & 198 & Coupling \\
        \hline
        \cross & \cross & \cross & 169 & Positioning \\
        \hline
        \cross & \cellcolor{yellow!30} \tick & \cross & 73 & \multirow{2}{*}{Tooling}\\
         \tick & \cellcolor{yellow!30} \tick & \cross & 64 & \\
        \hline
        \cross & \cross & \cellcolor{yellow!30}\tick & 44 & \multirow{3}{*}{Agitating}\\
        \cross & \tick &\cellcolor{yellow!30} \tick & 13 & \\
        \tick & \cross & \cellcolor{yellow!30}\tick & 1 & \\
        \hline
        \tick & \tick & \tick & 0 &  \\
        \hline
    \end{tabular}
    \caption{Feature-based characterization of annotated laboratory actions. Highlighted cells mark the dominant feature combinations that define the four taxonomic categories, while the final row shows that no actions were observed with all three features present.}
    \label{tab:features}
\end{table}

\kefi{
To move from individual annotated actions toward a structured taxonomy, each action was characterized using the three key features introduced in Section~\ref{sec:features}: whether it was force dependent, whether it involved indirect control, and whether it was realized through repetitive motion. Each annotated action in the dataset was marked against these features, producing a systematic representation of its execution profile.

From these profiles, and guided by the semantic intent of the actions, four overarching categories were distilled to capture the essential modes of laboratory manipulation:
\begin{itemize}
    \item Positioning: non–force-dependent, one-off operations under direct control.
    \item Coupling: force-dependent, one-off operations used to secure or release attachments.
    \item Agitating: actions characterized by repetitive or cyclic motion.
    \item Tooling: operations that manipulate substances indirectly by controlling a laboratory tool.
\end{itemize}

The results of the feature-based mapping are summarized in Table~\ref{tab:features}, which shows how the annotated actions cluster according to their execution requirements. Highlighted cells indicate the dominant feature that correspond to each category. For instance, force-dependent but non-cyclic operations align with coupling tasks such as fastening or releasing attachments, while cyclic operations map onto agitating tasks such as stirring or swirling. Positioning tasks, by contrast, lack all three features, reflecting their direct, one-off nature.

The table also reveals less intuitive combinations, which nonetheless fit consistently within the taxonomy. For example, actions that are both indirect and force-dependent, such as squirting, fall under tooling: although force is essential for control, the defining characteristic is that the operation is mediated through a tool for example using a syringe to inject liquids. Similarly, indirect cyclic actions such as stirring or swirling are grouped under agitating, while cyclic, force-dependent actions like grinding also fall into this category. The final row, which would correspond to actions that are simultaneously indirect, force-dependent, and cyclic, remains empty, as no such cases were observed.

Within each category, sub-actions are further distinguished along dimensions such as translational versus rotational motion, with additional individual actions defined to capture the diversity of laboratory practice. Collectively, these categories constitute TARMAC—a \underline{ta}xonomy for \underline{r}obotic \underline{ma}nipulation in \underline{c}hemistry—which provides a structured action space linking experimental intent to robotic execution. The details of the taxonomy are as follows. 
}

\subsection{Positioning}
Positioning refers to the controlled placement of objects by overcoming minor resistive forces such as gravity or friction to achieve a desired pose.  
Four main subtypes are distinguished. 
\begin{itemize}
    \item \emph{Transitional positioning} involves placing an object without substantially altering its orientation, as in setting a beaker on a benchtop or placing a weigh boat.  
    \item \emph{Rotational positioning} adjusts the orientation of an object during placement, for example when rotating a vial to align its label.  
    \item \emph{Insertive positioning} requires introducing an object into a constrained space, such as fitting a filter into a flask neck or docking a condenser.  
    \item \emph{Sliding positioning} maintains surface contact while moving an object to its target location, for example when closing a lid to ensure a proper seal.
\end{itemize}

\subsection{Coupling}
Coupling describes the joining or separation of objects through the application of controlled force to overcome resistance, friction, or material deformation, often using force measurements as endpoint signals.  
\begin{itemize}
    \item \emph{Transitional coupling} involves direct force along a linear path, such as pushing to install a tight stopper.  
    \item \emph{Rotational coupling} relies on torque, as when screwing or unscrewing a cap. 
    \item \emph{Elastic coupling} temporarily deforms a material when operating—for instance, stretching a rubber tube to widen an opening for attachment.  
    \item \emph{Penetrative coupling} \kefi{is a process where sustained force is applied to a material until its structural failure occurs}, as in puncturing a septum with a needle.  
\end{itemize}

\subsection{Tooling}
Tooling encompasses manipulations performed using tools or indirect forces such as gravity or applied pressure, thereby avoiding direct hand–object contact.  
\begin{itemize}
    \item \emph{Scooping} \kefi{is the immersion of a tool through particulate or fluid to capture a defined volume, such as collecting a reagent with a spatula.}
    \item \emph{Spotting} deposits or transfers small material quantities through surface tension or capillary action, without tool–surface contact, for instance when applying a sample to chromatography paper.  
    \item \emph{Pouring} transfers liquid by tilting a vessel to regulate gravity-driven flow, such as decanting a solution from a beaker.
    \item \emph{Squirting} \kefi{uses a pressure difference to aspirate or eject fluid through an orifice, as performed by a pipette.}
\end{itemize}

\subsection{Agitating}
Agitating encompasses repetitive or periodic motions that promote mixing, homogenization, cleaning, or material modification.  
\begin{itemize}
    \item \emph{Shaking} involves vigorous back-and-forth or vertical movements to accelerate mixing, for instance when dissolving solids in a volumetric flask. 
    \item \emph{Swirling} provides a gentler alternative, using circular motion to distribute material, such as coating the interior of a vial with solution. 
    \item \emph{Stirring} mixes liquids or suspensions by moving a rod or similar implement through the medium. 
    \item \emph{Twisting} applies axial rotation to manipulate material flow, for example when distributing grease between glassware joints.
    \item \emph{Swabbing} consists of repeated tool–surface contact to apply or remove material in confined areas, as in cleaning the neck of a flask whereas 
    \item \emph{Wiping} achieves broader coverage by drawing a tool across a surface, such as cleaning a pipette tip or blotting liquid from a slide.  
    \item \emph{Wrapping} secures materials by encircling them with a flexible medium, such as sealing a flask with parafilm. 
    \item \emph{Grinding} reduces particle size or texture through abrasive contact, as in crushing solids with a mortar and pestle. 
\end{itemize}

\section{Analysis and Discussion} \label{sec:analysis}

\begin{figure*}[t]
    \centering
    \includegraphics[width=\linewidth]{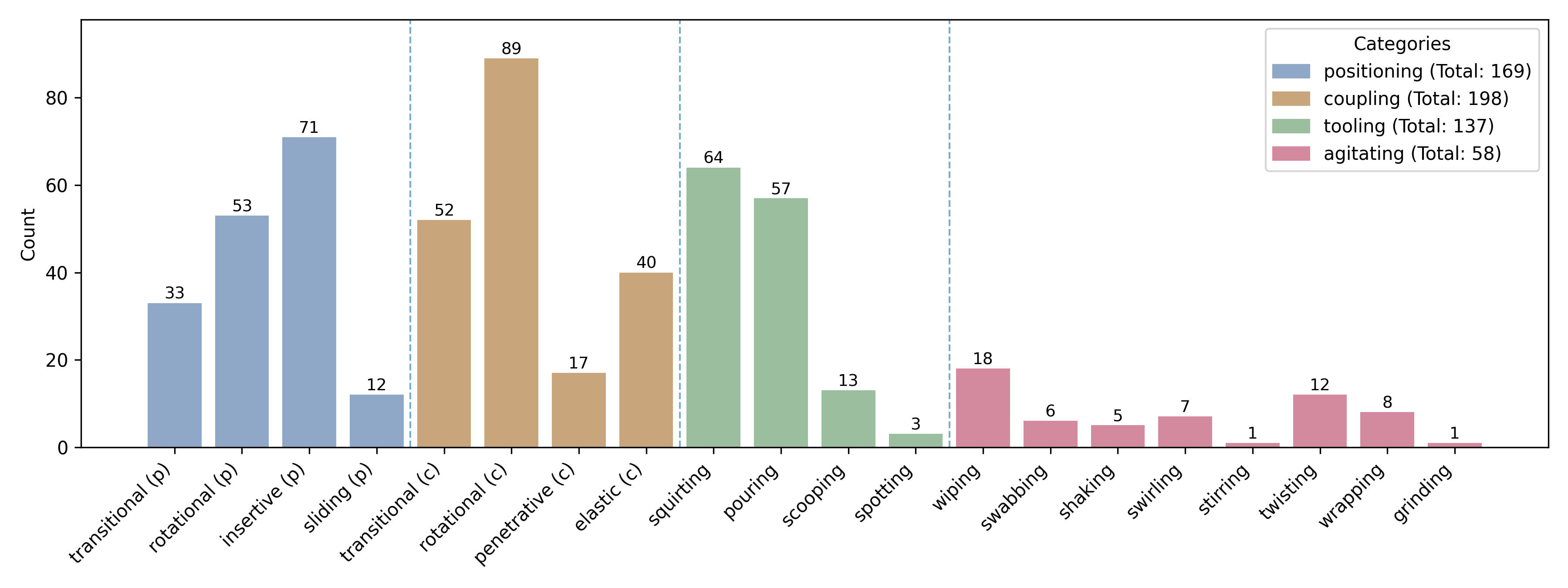}
    \caption{Distribution of taxonomic action labels across annotated chemistry teaching videos. Each bar represents the frequency of actions within the four major categories of TARMAC: positioning (blue), coupling (brown), tooling (green), and agitating (red). The relatively balanced counts across positioning, coupling, and tooling indicate comprehensive coverage of core manipulations, while the lower frequency of agitating actions reflects their more limited but specialized role in laboratory practice. This distribution highlights both the representativeness of the dataset and the discriminative structure of the taxonomy.}
    \label{fig:label_count}
\end{figure*}

\kefi{
This section evaluates the proposed taxonomy through two complementary studies. First, laboratory videos were re-annotated to build a dual natural-language and taxonomic dataset, illustrating how protocols can be linked to robot-executable skills. Second, complete laboratory tasks were decomposed into taxonomic primitives, demonstrating that the taxonomy provides sufficient coverage and granularity for practical application.
}

\subsection{Chemistry Manipulation Dataset}

To further evaluate the proposed taxonomy, chemistry laboratory videos and teaching materials were re-annotated using both the taxonomy and the original natural-language descriptions. The original annotations provide narrative descriptions such as “pipette solvent”, whereas the taxonomic relabeling maps these into structured action primitives such as “squirting”, thereby capturing the underlying execution policy. Each annotated instance comprises the natural language description, the corresponding taxonomic label, the associated video frame, and the full video sequence (see Figure~\ref{fig:chemistry_manipulation_dataset}). This dual annotation, \kefi{together with the feature profiles}, produces a dataset that is both human-interpretable and machine-actionable, facilitating downstream applications in benchmarking, imitation learning, and language-to-action translation.

The distribution of labels across the taxonomy was further analyzed, as shown in Figure~\ref{fig:label_count}. The results indicate that the categories of positioning, coupling, and tooling are relatively balanced, whereas agitating is comparatively underrepresented. This distribution highlights the discriminative power of the taxonomy: the near-even allocation across the major categories suggests a well-structured division of manipulation skills, while the lower frequency of agitating actions reflects their more limited role in laboratory procedures.

\begin{table*}[h]
    \centering
    \begin{tabular}{|p{0.5\linewidth}|p{0.15\linewidth} | p{0.25\linewidth}|}
        \hline
        \rowcolor{gray!20} \multicolumn{3}{|l|}{\textbf{Experimnent}: Hydrolysis of a nitrile} \\ \hline
        \multicolumn{3}{|c|}{%
            \rule{0pt}{3ex}
            \schemestart
                \chemfig{[:-60]*6(=(-NO_2)-=-(-[:30]-[:-30]CN)=-=)}
                \hspace{3em}
                \arrow{->[\ce{H2SO4}][\ce{H2O}, $\Delta$, 30 min]}
                \hspace{3em}
                \chemfig{[:-60]*6(=(-NO_2)-=-(-[:30]-[:-30](=O)-[:-90]OH)=-=)}
            \schemestop
            \rule[-6ex]{0pt}{0pt}
        } \\ 
        \multicolumn{3}{|p{0.9\linewidth}|}{A 9 M solution of sulfuric acid in water was carefully prepared in a conical flask. A 100 mL single-neck round-bottomed flask with a stirrer bar was charged with 4-nitrophenylacetonitrile followed by 9 M sulfuric acid (25 mL) and the suspension was refluxed for 30 minutes. After this time, the solution was allowed to cool to room temperature before the addition of water (25 mL) and cooling in an ice-bath. The crude product was collected via vacuum filtration.
The crude solid was dissolved in ethyl acetate (20 mL) and extracted with 2 M sodium hydroxide solution (20 mL). The two layers were separated, and the aqueous layer was acidified with 3 M hydrochloric acid (20 mL) and extracted with ethyl acetate (20 mL). The organic layer was dried over magnesium sulfate, filtered and concentrated to yield the product which was recrystallised from 20\% ethanol in water.
} \\ \hline
         \rowcolor{gray!20} \textbf{Action needed to conduct experiments}&\textbf{Taxonomy Category} & \textbf{Taxonomy action} \\ \hline
        A magnetic stirrer bar was placed in round bottom flask (RBF) & Positioning & Insertive positioning \\ \hline
        0.486 g 4-nitrophenylacetonitrile was weighed onto weighing boat & Tooling & Scooping \\ \hline
        Weighing boat was tipped into RBF & Tooling & Pouring \\ \hline
        25 mL of sulfuric acid solution was added to RBF & Tooling & Pouring \\ \hline
        Condenser added to the top of RBF & Positioning & Insertive positioning \\ \hline
        Suspension was heated to reflux for 30 minutes. && \\ \hline
        Condenser was removed & Positioning & Transitional positioning \\ \hline
        25 mL water added. & Tooling & Pouring \\ \hline
        RBF placed an ice-bath for cooling. & Positioning & Transitional positioning \\ \hline
        Sinter funnel placed on top of Büchner flask and rubber bung. & Positioning & Insertive positioning \\ \hline
        Contents of RBF poured into sinter funnel. & Tooling & Pouring \\ \hline
        Solid collected scraped into RBF. & Tooling & Scooping \\ \hline
        20 mL ethyl acetate added to RBF. & Tooling & Pouring \\ \hline
        Solution poured into separating funnel. & Tooling & Pouring \\ \hline
        20 mL sodium hydroxide solution (2 M) added to separating funnel. & Tooling & Pouring \\ \hline
        Stopper placed on top of separating funnel. & Coupling & Force Coupling \\ \hline
        Separating funnel shaken. & Agitating & Shaking \\ \hline
        Stopper removed from separating funnel. & Coupling &  Force Coupling \\ \hline
        Beaker placed underneath separating funnel. & Positioning & Transitional positioning \\ \hline
        Tap of separating funnel opened. & Positioning & Rotational Positioning \\ \hline
        New beaker placed underneath separating funnel. & Positioning & Static positioning \\ \hline
        Tap of separating funnel opened. & Positioning & Rotational Positioning \\ \hline
        20 mL hydrochloric acid (3 M) added to 1st measuring beaker. & Tooling & Pouring \\ \hline
        Mixture from measuring beaker added to separating funnel. & Tooling & Pouring \\ \hline
        20 mL ethyl acetate added to separating funnel. & Tooling & Pouring \\ \hline
        Stopper placed on top of separating funnel. & Coupling & Force Coupling \\ \hline
        Separating funnel shaken. & Agitating & Shaking \\ \hline
        Stopper removed from separating funnel. & Coupling & Force Coupling \\ \hline
        Beaker placed underneath separating funnel. & Positioning & Transitional positioning \\ \hline
        Tap of separating funnel opened to remove 1st layer. & Positioning & Rotational Positioning \\ \hline
        New beaker placed underneath separating funnel. & Positioning & Transitional positioning \\ \hline
        Tap of separating funnel opened to remove 2nd layer. & Positioning & Rotational Positioning \\ \hline
        Beaker placed underneath separating funnel removed. & Positioning & Transitional positioning \\ \hline
        Magnesium sulfate added to beaker to reach ‘snow globe effect’. & Tooling & Scooping \\ \hline
        Funnel placed on top of new RBF. & Positioning & Insertive positioning \\ \hline
        Contents of beaker poured into sinter funnel. & Tooling & Pouring \\ \hline
        Funnel removed from RBF. & Positioning & Transitional positioning \\ \hline
        RBF attached to rotary evaporator to remove solvent. & Positioning & Insertive Positioning \\ \hline
        Product from RBF scraped into conical flask. & Tooling & Scooping \\ \hline
        Conical flask placed on hotplate. & Positioning & Static positioning \\ \hline
        20\% ethanol in water mixture added dropwise to conical flask. & Tooling & Squirting \\ \hline
        Solid dissolved in the flask. & Agitating & Swirling \\ \hline
    \end{tabular}
    \caption{Task decomposition of the hydrolysis of a nitrile experiment. The high-level laboratory protocol is expanded into the concrete manipulations a chemist would perform (left), which are then mapped to structured TARMAC action primitives (right), showing how implicit laboratory practice can be systematically expressed through the taxonomy.}
    \label{tab:task_decomposition}
\end{table*}

\subsection{Task Decomposition}

To evaluate the comprehensiveness of the taxonomy, a task decomposition study was conducted on undergraduate laboratory experiments. Starting from the brief instructions provided in the laboratory manual, a chemist expanded each step into the concrete actions they would actually perform in practice. These were then translated into taxonomic primitives, producing a structured breakdown of the experiment. An example is shown in Table~\ref{tab:task_decomposition}, where the hydrolysis of a nitrile is decomposed from human-oriented instructions into taxonomic actions.

This exercise demonstrates that the taxonomy, originally derived from teaching-lab videos, provides sufficient coverage to represent essential operations required in typical laboratory experiments. By grounding protocol instructions in explicit taxonomic actions, the taxonomy is shown to capture both the breadth and granularity of skills required for real laboratory practice, offering a comprehensive foundation for robotic execution.

\section{Preliminary Experiment Validation} \label{sec:validation}

The taxonomy introduced in Section~\ref{sec:taxonomy} provides a structured description of laboratory manipulations. To move beyond its role as a descriptive scheme, its applicability to practical robotic execution was examined. For this purpose, a preliminary framework was developed in which the taxonomy was instantiated as executable action primitives. Within this framework, targeted experiments could be planned and executed, and workflows constructed directly from taxonomic actions—thereby validating the taxonomy’s utility in real-world laboratory contexts.

\subsection{Taxonomy-Driven Framework}

Our validation framework is designed to demonstrate how the proposed taxonomy can move beyond a descriptive role and actively support robotic execution in laboratory environments. The framework is guided by two requirements. First, chemists need an interface that is natural and intuitive, without requiring programming expertise. Second, robotic systems must remain flexible, adapting to the variability of experimental conditions and evolving laboratory needs.

Within this framework, the taxonomy provides the structure that connects human intent with robotic capability. \kefi{The taxonomy is instantiated as a library of action primitives, ensuring a direct one-to-one mapping between each taxonomic skill and a minimal robot-executable operation (e.g., the taxonomic action "pouring" is executed by the pour action primitive).} Frequently recurring combinations of primitives are further encapsulated into a \textit{macro}, offering reusable higher-level commands. By exposing both primitives and macros through a standard interface, the framework allows natural language task specifications to be translated into concrete robotic actions.

To realize this translation, the recently proposed \textit{Model Context Protocol (MCP)} \cite{anthropicmcp} was adopted, which standardizes the way large language models (LLMs) and other reasoning models interact with external tools. In the implementation, every primitive and macro derived from the taxonomy was registered as an MCP tool.
This allows a user instruction—such as “prepare a solution”—to be decomposed into a sequence of taxonomic actions, executed step by step while incorporating feedback. The MCP interface also enables dynamic modification of the available skill set: new macros can be registered during an experiment, or specific tools can be disabled to enforce safety constraints.

Beyond robotic manipulation, the framework remains open to integration with other laboratory automation modules. External services—such as existing robotic platforms, scheduling software, or sensing systems—can be exposed as MCP tools alongside the taxonomy-driven actions, enabling seamless coordination. In this way, the taxonomy serves as the backbone of the framework, grounding abstract protocol descriptions in a structured and extensible library of robot-executable skills while retaining flexibility and safety through standardized tool interaction.

\subsection{Action Primitives and Macros}

A central feature of this framework is the instantiation of the taxonomy into \textit{action primitives}. Each primitive corresponds to a minimal, parameterized skill representing a distinct laboratory operation (e.g., pouring, stirring, swirling). These primitives capture the fine granularity of manipulations and serve as the fundamental building blocks from which more complex experimental routines can be assembled. Importantly, the taxonomy does not prescribe how a primitive must be realized: a single unit may be implemented using motion planning pipelines, rule-based controllers, or modern learning-based approaches, depending on the laboratory setup. This flexibility ensures that the taxonomy can be grounded in heterogeneous robotic systems while maintaining a common representational layer.

Analysis of laboratory procedures indicates that many operations recur in predictable patterns. For example, “pick up beaker $\rightarrow$ pour contents $\rightarrow$ mix” appears across multiple solution preparation protocols. To capture such regularities, \textit{macros} are introduced as reusable higher-level commands constructed from compositions of primitives. Macros abstract away low-level details that are often implicit in laboratory practice (e.g., clamping a vessel before heating) while encapsulating common experimental routines into modular, reusable workflows. In this way, macros reduce cognitive load, simplify task specification, and enable scalable automation without sacrificing the ability to revert to fine-grained primitives when precision is required.

Macros can be derived in several ways. The most direct approach is manual composition, where developers or lab technicians explicitly combine primitives into reusable workflows. Beyond this, more accessible methods can make macro creation faster and more intuitive. For example, LLM-based synthesis can exploit the model’s coding and planning abilities, expanding a chemist’s natural-language instruction into a structured sequence of primitives. Graphical programming interfaces such as Blockly or Scratch provide a drag-and-drop environment where primitives are assembled visually, lowering the barrier for non-programmers. Finally, vision-language models may infer common routines directly from video demonstrations, suggesting macros grounded in real laboratory practice.
Both manual definition and LLM-based synthesis were explored in the implementation, but the broader point is that grounding macros in taxonomic actions enables diverse methods of creation, ensuring adaptability across different users, contexts, and levels of expertise.

Together, primitives and macros form a layered skill library grounded in the taxonomy. Primitives ensure coverage of the essential operations that define laboratory practice, while macros provide scalable abstractions that align with experimental routines. This dual structure enables flexible task execution across levels of granularity, supporting both precise manipulation and efficient workflow automation.

\begin{figure}[h]
    \centering
    \includegraphics[width=\linewidth]{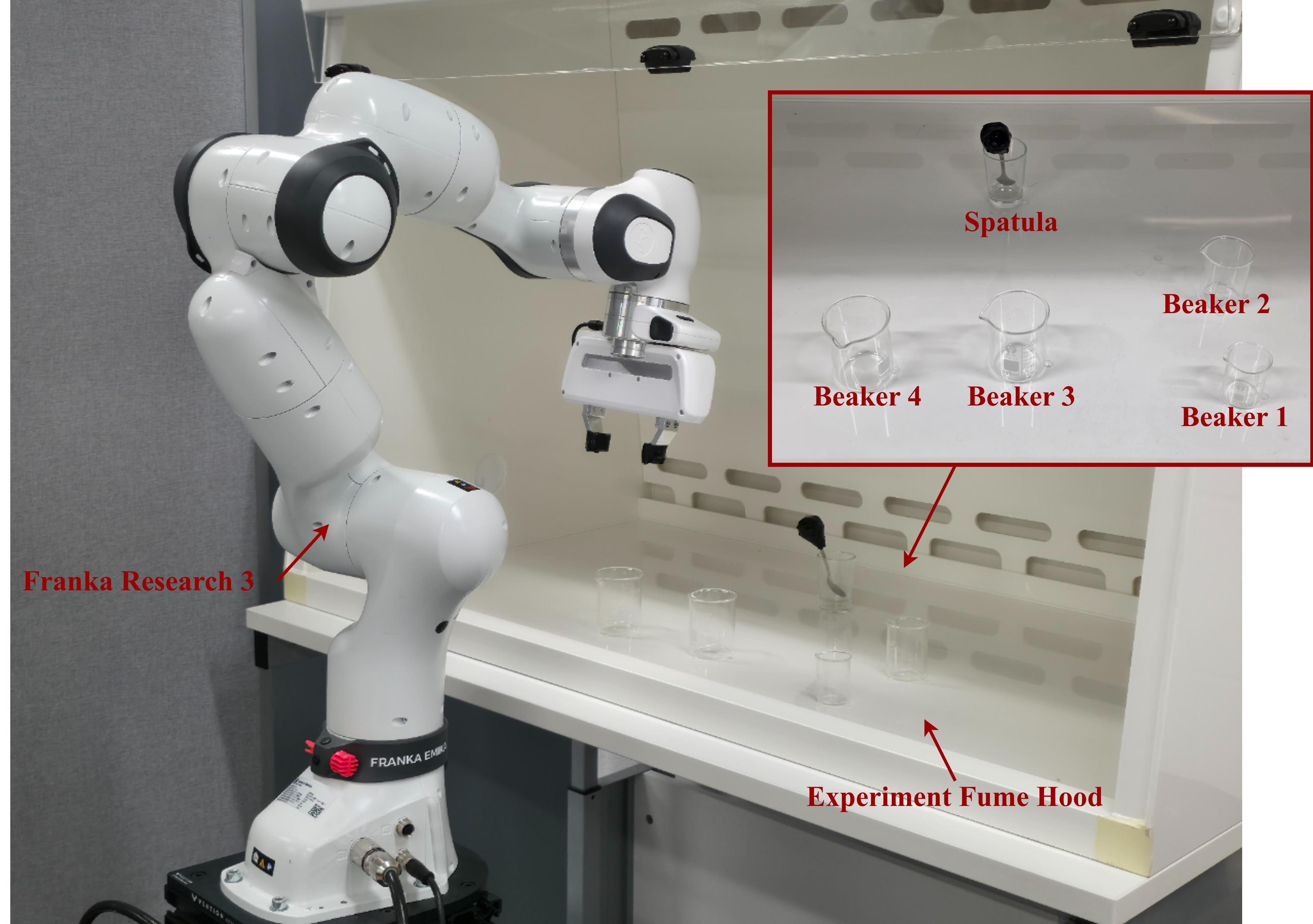}
    \caption{Experimental setup for taxonomy-driven validation. A Franka Research 3 robotic arm, positioned beside a standard laboratory fume hood, is configured to perform representative tasks. The workspace includes beakers pre-filled with solvent and solute, along with a spatula used for stirring.}
    \label{fig:llm_setup}
\end{figure}

\begin{figure}[p]
    \centering
    \includegraphics[width=\linewidth]{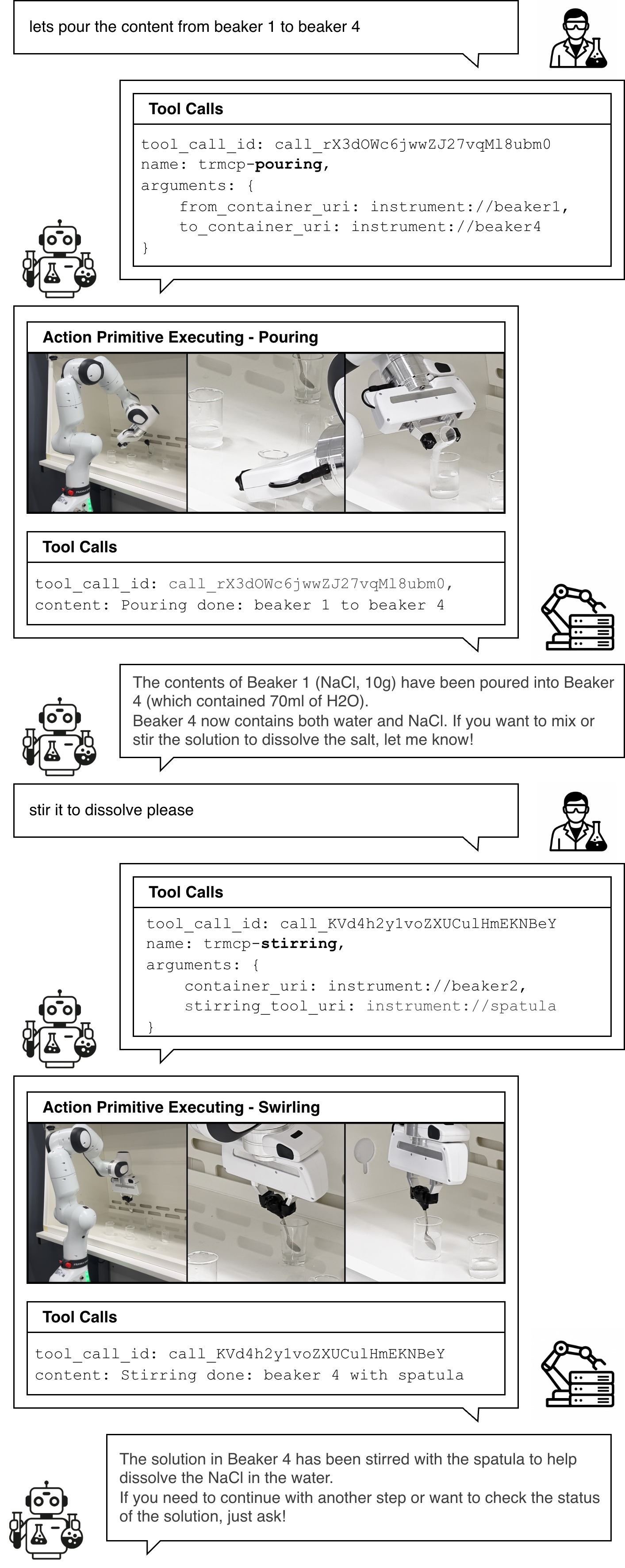}
    \caption{Example execution trace of taxonomy-driven interaction with a large language model (LLM). The model receives a high-level natural-language instruction, decomposes it into TARMAC primitives, and issues sequential tool calls.}
    \label{fig:llm_conversation_one}
\end{figure}

\subsection{Experiment Setup}
To validate the feasibility of the taxonomy-driven framework, a series of proof-of-concept experiments was designed in which representative primitives were implemented and applied to complete a simple but meaningful laboratory procedure: preparing a sodium chloride solution. The experiments had two objectives: (1) to verify that primitives instantiated from the taxonomy can be reliably executed on a robotic platform, and (2) to demonstrate that higher-level workflows can be composed from these primitives. Together, these tasks provide an initial validation of the taxonomy and show its practical value in connecting protocol-level instructions with robot-executable behaviors.

\subsubsection{System Configuration}

All experiments were conducted using a Franka Research 3 robotic arm positioned beside a standard laboratory fume hood. The workspace was arranged with four beakers, each pre-assigned with contents (e.g., water or sodium chloride powder), and a spatula used as the stirring tool. Figure~\ref{fig:llm_setup} illustrates the experimental setup.

The robot was connected to an MCP server that exposed a subset of action primitives representative of common laboratory operations. For this proof-of-concept, three primitives were implemented:
\begin{itemize}
    \item Pouring: transfers liquid or powder contents from a source beaker to a destination beaker (two parameters).
	\item Stirring: agitates the contents of a beaker using a specified stirring tool (two parameters).
	\item Swirling: agitates the contents of a beaker by circular motion without external tools (one parameter).
\end{itemize}

These primitives were selected because they are both frequent in chemistry protocols and illustrative of two categories in this taxonomy: \textit{tooling} (pouring) and \textit{agitating} (stirring, swirling). Each primitive was implemented as an asynchronous Python function that invokes a motion planning pipeline to generate collision-free trajectories for the robotic arm. Input parameters and return values were explicitly defined; for example, the \texttt{pouring} primitive accepts the  \kefi{identifiers} of the source and destination beakers and returns the URI of the destination beaker upon completion.

\subsubsection{Task Design}

Two different tasks were designed, both centered on preparing a sodium chloride solution -- a canonical laboratory procedure involving the combination of a solvent (water) with a solute (sodium chloride powder) followed by mixing. This task was selected because it is both representative of real laboratory practice and sufficiently simple to allow controlled evaluation.  

A foundation large language model (LLM), specifically GPT-4.1 used without task-specific fine-tuning or domain adaptation, served as the decision-making agent. The LLM interacted with the MCP server, which exposed the action primitives as callable tools, thereby grounding high-level instructions in executable robot actions.  

\textbf{Direct Task Execution}: In this task, the LLM was instructed only to prepare a sodium chloride solution. No guidance was given on which primitives to use. The experiment evaluated whether the model could autonomously select the appropriate beakers, invoke the relevant primitives (e.g., \texttt{Pouring}, \texttt{Stirring}/\texttt{Swirling}), and maintain the correct sequence of operations to complete the preparation.
    
\textbf{Action Macros Composition}: This task examined whether the LLM could generate a new action macro from existing primitives. Instead of invoking \texttt{Pouring} and then \texttt{Stirring} separately each time, the model was instructed to create a higher-level tool (e.g., \texttt{PrepareSolution}) that encapsulates the sequence. 

\begin{figure*}[p]
\centering
\begin{minipage}{\textwidth}
\begin{lstlisting}[language=Python, frame=single, caption={Code snippet of an action macro automatically generated by the LLM \kefi{with instruction prompt \textit{"lets create a new tool such that it takes two beakers as input, if one is empty, use the other one, to pour into a target beaker. and then do swirling and string to mix. finally return what beaker has been used if success."}. The macro encapsulates this pour-and-mix sequence into a reusable higher-level command.}}, label={lst:pour_and_stir_solution}]
async def beaker_pour_mix(
    beaker_a_uri: str, beaker_b_uri: str, target_beaker_uri: str, stirring_tool_uri: str
) -> str:
    """
    Pour from the first non-empty of two source beakers into a target beaker, then swirl and stir to mix.

    Args:
        beaker_a_uri: URI of the first source beaker.
        beaker_b_uri: URI of the second source beaker.
        target_beaker_uri: URI of the target beaker to receive contents.
        stirring_tool_uri: URI of the stirring tool.

    Returns:
        The URI of the beaker that was poured from.

    Raises:
        ValueError: If any URI is not a non-empty string.
        Exception: If pouring from both beakers fails (e.g., both empty or not found).
    """
    for v in (beaker_a_uri, beaker_b_uri, target_beaker_uri, stirring_tool_uri):
        if not isinstance(v, str) or not v.strip():
            raise ValueError("All URIs must be non-empty strings.")
    try:
        await provider.pouring(beaker_a_uri, target_beaker_uri)
        used = beaker_a_uri
    except Exception:
        try:
            await provider.pouring(beaker_b_uri, target_beaker_uri)
            used = beaker_b_uri
        except Exception as e:
            raise e
    await provider.swirling(target_beaker_uri)
    await provider.stirring(target_beaker_uri, stirring_tool_uri)
    return used
\end{lstlisting}
\end{minipage}

\begin{minipage}{\textwidth}
\begin{subfigure}{0.32\textwidth}
    \centering
    \includegraphics[width=\linewidth]{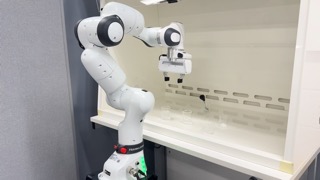}
    \caption{Initial Position}
\end{subfigure}
\begin{subfigure}{0.32\textwidth}
    \centering
    \includegraphics[width=\linewidth]{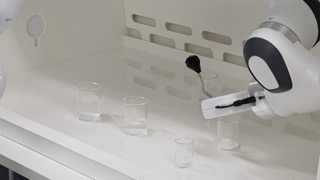}
    \caption{Pick correct non-empty beaker}
\end{subfigure}
\begin{subfigure}{0.32\textwidth}
    \centering
    \includegraphics[width=\linewidth]{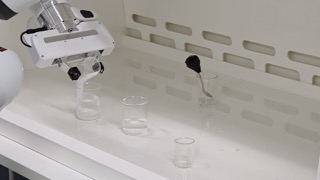}
    \caption{Pouring}
\end{subfigure}
\end{minipage}

\begin{minipage}{\textwidth}
\begin{subfigure}{0.32\textwidth}
    \centering
    \includegraphics[width=\linewidth]{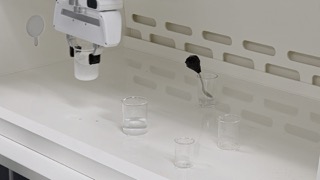}
    \caption{Swirling}
\end{subfigure}
\begin{subfigure}{0.32\textwidth}
    \centering
    \includegraphics[width=\linewidth]{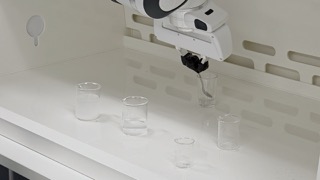}
    \caption{Pick up spatula}
\end{subfigure}
\begin{subfigure}{0.32\textwidth}
    \centering
    \includegraphics[width=\linewidth]{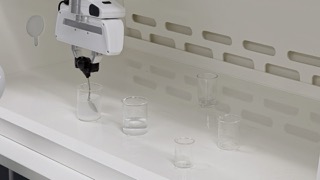}
    \caption{Stirring}
\end{subfigure}
\end{minipage}
\caption{Illustration of macro composition and execution. (a–f) Execution of the macro on the robotic platform, showing the robot preparing a sodium chloride solution by transferring solute into solvent and mixing. This demonstrates how TARMAC primitives can be abstracted into macros, enabling scalable automation of laboratory workflows.}
\label{fig:llm_workflow}

\end{figure*}
    
\subsection{Analysis of Results}

The two validation tasks collectively demonstrate the feasibility of the proposed taxonomy-driven framework. Each task highlights a different capability of the system: the basic executability of primitives and the compositionality of macros. Full execution traces and logs are provided in the supplementary material.

The \textbf{Direct Invocation} task confirmed that individual action primitives can be reliably executed in sequence. The agent correctly selected the relevant beakers, invoked the \texttt{pouring} primitive to transfer solvent into solute, and followed with the instructed mixing operation. The execution record (Figure~\ref{fig:llm_conversation_one}) shows that the correct sequence was composed without additional guidance, validating that primitives grounded in the taxonomy can be effectively mapped from high-level instructions to low-level robotic actions.

The \textbf{Macro Composition} task demonstrated that the framework supports dynamic abstraction. 
From prior observations, the agent generated a reusable macro, \texttt{beaker\_pour\_mix}, encapsulating the sequence of pouring followed by stirring and swirling to mix. 
The automatically generated code (Code Example~\ref{lst:pour_and_stir_solution}) included parameter validation and error handling, reflecting an understanding of both functional and safety requirements. 
Once registered, the macro was successfully invoked to prepare a solution (Figure~\ref{fig:llm_workflow}), confirming that higher-level routines can be synthesized from primitives and reused as building blocks across tasks.

Together, these results demonstrate the practical value of the taxonomy as a bridge between protocol-level instructions and robot-executable behaviors. By instantiating manipulation categories as primitives and supporting their composition into macros, the taxonomy provides both coverage of laboratory manipulations and a scalable path toward reusable automation modules. This proof-of-concept establishes a foundation for extending the primitive library and formalizing macro collections in future work.

\kefi{
\section{Challenges and Outlook} \label{sec:challenges}

Our study highlights several challenges in bridging human laboratory practice with robotic execution. A first observation comes from comparing task decompositions with video-based annotations. While human-written protocols typically provide only high-level instructions, teaching-lab videos reveal many fine-grained actions that are essential for successful execution. These implicit steps -- such as securing glassware before use -- are trivial for human practitioners but must be explicitly modeled for robots. This under-specification represents a fundamental gap between human-oriented laboratory documentation and the requirements of robotic autonomy.

A second challenge concerns the inherent difficulty of certain laboratory manipulations. For instance, elastic coupling, common in air-sensitive experiments, requires tight sealing with rubber components. These actions are challenging not only because they demand precise control in confined spaces, but also because the lack of accurate models makes the behavior of elastic materials unpredictable. In practice, elastic components often need to be stretched or expanded before insertion, adding another layer of complexity. Similarly, squirting actions, such as those performed with pipettes, are more demanding than pouring because they require precise yet indirect control over the velocity of liquid ejection, in addition to interacting with elastic elements such as pipette tips. Finally, some categories of actions are infrequent yet equally challenging. For example, wrapping delicate glassware demands both secure handling and extreme care to avoid breakage —- skills that remain well beyond current robotic dexterity.

Looking forward, these challenges also highlight promising directions. The taxonomy offers a framework for making implicit knowledge explicit, thereby supporting the construction of standardized datasets and benchmarks that better reflect real laboratory practice. Advances in robotic hardware, particularly dexterous end-effectors , force torque sensors or tactile sensors, will be crucial to tackling elastic and delicate manipulations. Ultimately, the taxonomy serves not only as a descriptive tool but also as a stepping stone toward robotic chemists capable of executing complex workflows with reliability and precision. By identifying current gaps and charting pathways for addressing them, this work lays the groundwork for future systems that combine structured representations, advanced perception, and adaptive control to achieve higher levels of autonomy in the laboratory.
}

\section{Conclusion} \label{sec:conclusion}
This work introduced TARMAC, a taxonomy for robot manipulation in chemistry, designed to capture the core skills underpinning laboratory practice. Derived from teaching-lab demonstrations and validated through force–torque experiments, TARMAC provides a structured vocabulary that organizes manipulations by their functional role and execution requirements. At the same time, the analysis highlights challenges that remain, particularly in handling elastic, delicate, and velocity-dependent manipulations that demand more advanced sensing and control.

Beyond its descriptive value, the taxonomy can be instantiated as robot-executable primitives and composed into reusable macros, thereby bridging protocol-level descriptions familiar to chemists with the concrete actions required by robotic systems. Preliminary validation demonstrates that TARMAC supports both direct execution of primitives and the construction of higher-level workflows, enabling systematic reuse of skills and scalable automation.

Taken together, these contributions establish TARMAC as a foundation for more interpretable, extensible, and autonomous laboratory automation. Just as tarmac in the physical world provides the structured surface upon which vehicles travel, this taxonomy serves as a common ground on which robotic chemists can operate. By making implicit skills explicit, it opens new opportunities for benchmarking, imitation learning, and integration with reasoning models such as LLMs. Ultimately, taxonomy-driven approaches like TARMAC have the potential to accelerate progress toward reliable robotic chemists capable of executing complex experimental workflows with safety, precision, and adaptability.

\bibliographystyle{IEEEtran}
\bibliography{ref}
\todo{a few more citations}





\end{document}